\documentclass[10pt,twocolumn,letterpaper]{article}

\usepackage[pagenumbers]{iccv} % To force page numbers, e.g. for an arXiv version
\usepackage{array}
\usepackage{float}
\usepackage{multirow}

\usepackage{graphicx}
\usepackage{bbding}
\usepackage{setspace}
\usepackage[ruled,linesnumbered]{algorithm2e}  
\usepackage{amsmath} 

\PassOptionsToPackage{normalem}{ulem}
\usepackage{ulem}
\providecommand{\tabularnewline}{\\}
\floatstyle{ruled}
\newfloat{algorithm}{tbp}{loa}
\providecommand{\algorithmname}{Algorithm}
\floatname{algorithm}{\protect\algorithmname}
\definecolor{iccvblue}{rgb}{0.21,0.49,0.74}
\usepackage[pagebackref,breaklinks,colorlinks,allcolors=iccvblue]{hyperref}

\def\paperID{13862} % *** Enter the Paper ID here
\def\confName{ICCV}
\def\confYear{2025}

\title{PLOOD:  Partial Label Learning with Out-of-distribution Objects}

\author{Jintao Huang\\
Hong Kong Baptist University\\
{\tt\small jthuang@hkbu.edu.hk}
\and
Yiu-Ming Cheung\\
Hong Kong Baptist University\\
{\tt\small ymc@comp.hkbu.edu.hk}
\and
Chi-Man Vong\\
University of Macau\\
{\tt\small cmvong@um.edu.mo}
}

\begin{document}
\maketitle
\begin{abstract}
Existing Partial Label Learning (PLL) methods posit that training and test data adhere to the same distribution, a premise that frequently does not hold in practical application where Out-of-Distribution (OOD) objects are present. We introduce the OODPLL paradigm to tackle this significant yet underexplored issue. And our newly proposed PLOOD framework enables PLL to tackle OOD objects through Positive-Negative Sample Augmented (PNSA) feature learning and Partial Energy (PE)-based label refinement. The PNSA module enhances feature discrimination and OOD recognition by simulating in- and out-of-distribution instances, which employ structured positive and negative sample augmentation, in contrast to conventional PLL methods struggling to distinguish OOD samples. The PE scoring mechanism combines label confidence with energy-based uncertainty estimation, thereby reducing the impact of imprecise supervision and effectively achieving label disambiguation. Experimental results on CIFAR-10 and CIFAR-100, alongside various OOD datasets, demonstrate that conventional PLL methods exhibit substantial degradation in OOD scenarios, underscoring the necessity of incorporating OOD considerations in PLL approaches. Ablation studies show that PNSA feature learning and PE-based label refinement are necessary for PLOOD to work, offering a robust solution for open-set PLL problems.
\end{abstract}  

\section{Introduction}

Partial Label Learning (PLL) represents a weakly supervised learning framework where each instance is associated with a collection of candidate labels, of which only one is accurate. PLL markedly decreases labeling expenses by mitigating the necessity for exact annotations \cite{hu2023dual,zhong2024negative,xie2024class}. The main goal of PLL is to create multi-class classifiers that can accurately pick out the true label from a group of possible labels while being resistant to noise. PLL has gained significant popularity across various domains, including image annotation, web mining, and ecological informatics \cite{liu2023consistent, DBLP:conf/iccv/ZhangLZOTZ23, DBLP:conf/miccai/DongKV22, tian2024crosel}. This has emerged as a significant research domain within artificial intelligence \cite{song2020learning,si2024partial}.

Current PLL methods are effective in various scenarios; however, they rely on the closed-set assumption that the distributions of test and training data are identical. This assumption seldom holds in practical application. Out-of-Distribution (OOD) samples that significantly diverge from training distributions frequently occur in real-world applications \cite{pei2022out,albert2022embedding,zheng2024out,chen2024secure}. Conventional PLL models lack explicit mechanisms for detecting OOD instances, often resulting in their misclassification into predefined categories. This misclassification results in considerable performance decline and compromises model reliability and security. As shown in Figure 1, existing advanced PLL methods like PaPi \cite{xia2023towards}, CSDLE \cite{he2023candidate}, CroSel \cite{tian2024crosel}, SoDisam \cite{jiang2024navigating}, and LS-PLL \cite{gong2024does} are very accurate in closed-set (no-OOD) situations, but their accuracy drops by as much as 23.47±\% in open-set situations with OOD samples. This limitation underscores the pressing necessity to integrate OOD object detection into PLL, while it has yet been largely neglected.

\begin{figure}[tbh]
\centering \includegraphics[scale=0.65]{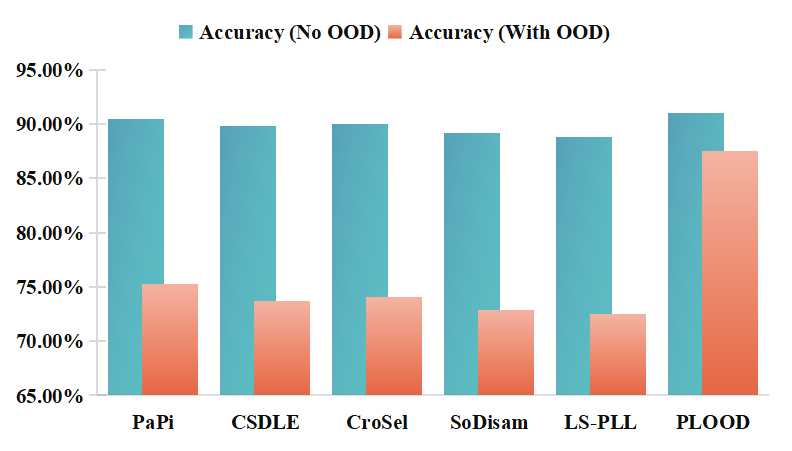} \caption{Practical example of SOTA PLL methods on CIFAR-10 under No- and With- OOD obejcts.}
\end{figure}
\begin{figure*}[tbh]
\centering \includegraphics[scale=0.36]{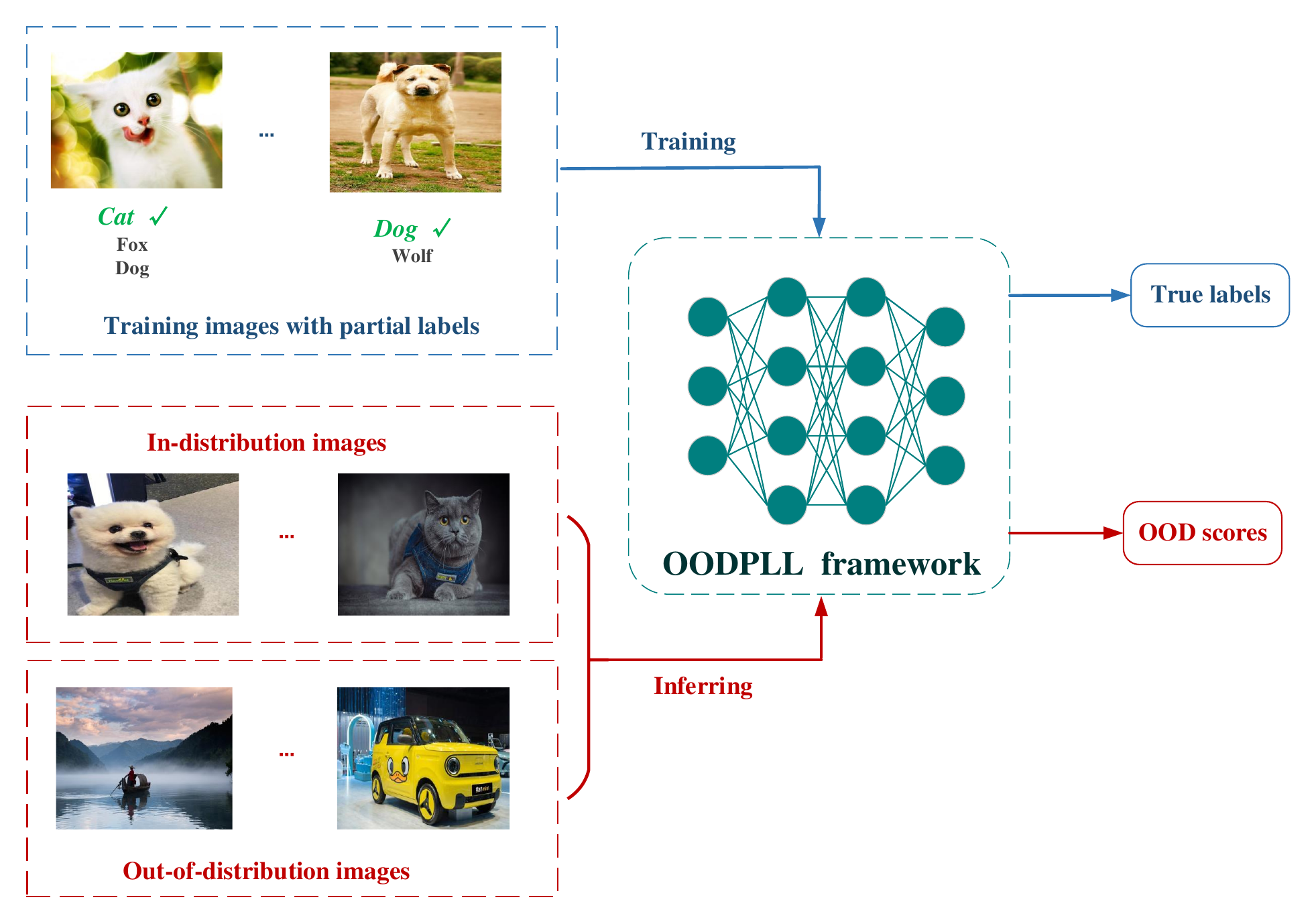} \caption{An illustration of OODPLL.  Top-left  is a scenario for training in animal image classification.  Each image possesses a collection of candidate labels, including Cat, Fox, and Dog; however, only Cat represents the accurate label.  Bottom-left are the images of in-distribution (ID) and out-of-distribution (OOD) animals.  In these instances, the OODPLL task must identify the true labels of the animals (ID objects) and detect the non-animal category (i.e., OOD objects).}
\end{figure*}

In response to these challenges, we introduce a new task termed Out-of-Distribution Partial Label Learning (OODPLL). Figure 2 illustrates a practical example of OODPLL, which involves the classification of animal images. The task of OODPLL involves accurately identifying the appropriate label for in-distribution (ID) samples while effectively distinguishing OOD samples that belong to entirely different categories. Integrating advanced OOD detection techniques into PLL appears promising; however, it presents two significant challenges: (1) Traditional PLL models exhibit a deficiency in learning features from OOD examples, resulting in their inability to distinguish between ID and OOD samples \cite{gong2021generalized,lv2023robustness,shi2023unreliable}.In PLL, OOD detection can be challenging because of label ambiguity that leads to wrong classification. False positives or negatives can happen if the weights of ID and ODD objects are not set correctly \cite{yang2021generalized,song2023recovering}. The challenges highlight the need for a new approach to effectively address OODPLL \cite{wang2021can,wang2022multi,yan2022ovae}.

This paper presents a novel framework namely Partial Label Learning with Out-of-Distribution Objects (PLOOD) to systematically address these issues of OODPLL. The composition consists of three interrelated innovations. Firstly, the Positive-Negative Sample Augmentation (PNSA) module improves feature representations by creating structured positive and negative augmentation samples building on positive-unlabeled learning (\cite{elkan2008learning,bekker2020learning}), which enables PLL to effectively detect the difference between ID and OOD instances. Secondly, a newly dynamic confidence calibration method for PLL is designed based on the robust feature representations from the PNSA, which can tackle the inherent ambiguity of labels by systematically enhancing the confidence estimates associated with them. Furthermore, we present a novel Partial Energy (PE) detecting score utilizing calibrated label confidences to leverage label uncertainty to energy-based modeling in a dynamic way. Thus, by utilizing the three innovative modules mentioned above, the proposed PLOOD may simultaneously properly predict labels for unseen ID samples while effectively detecting OOD objects in OODPLL scenarios.

\textbf{Contributions}: Here is a summary of this paper's primary contributions: (1) We present the OODPLL paradigm, emphasizing its importance in improving the robustness of PLL in open-set real-world scenarios. (2) We present the PLOOD framework, which uses structured PNSA to enable explicit OOD-aware feature learning, which can facilitate the differentiation of ID and OOD samples. (3) We introduce a novel PE scoring mechanism that effectively integrates label confidences with energy-based OOD detection, thereby addressing the challenges associated with ambiguous labeling. (4) Extensive evaluations on benchmark datasets, specifically CIFAR-10 and CIFAR-100 with several OOD data, indicate that current PLL methods exhibit significantly degraded performance in the presence of OOD data. Conversely, our PLOOD framework can maintain the accuracy of ID classification while effectively detecting OOD objects.

\textbf{PLOOD vs. Traditional PLL Methods}: Traditional PLL methods primarily concentrate on label disambiguation under a closed-set assumption \cite{lyu2022deep,DBLP:conf/cvpr/XiaL00G23,yan2024federated,tian2024crcd}, neglecting the existence of OOD samples. Consequently, when confronted with OOD instances, they misclassify these into pre-existing categories, leading to significant performance decline. PLOOD explicitly models OOD learning through PNSA augmentation, thereby preventing errors and ensuring robust classification.

\textbf{PLOOD vs. Existing OOD Detection Methods}: Recent advancements in Out-of-distribution Multi-label Learning (OODMLL) \cite{wang2021can, wang2022multi, zolfi2024yolood,zhang2023theoretical} have yielded encouraging outcomes. However, these methods presuppose a known exact label space, which is not applicable in PLL, where the true label is concealed among various candidates. The absence of exact supervision renders the direct adaptation of OODMLL methods ineffective. PLOOD addresses the challenge of OOD detection under partial label constraints by integrating PNSA-based feature learning with PE-based adaptive scoring.

\section{Proposed Methodology}

We assume a PLL training datasets with $N$ same distribution samples
$\mathbf{D}=\{(\mathbf{x}_{i},\mathbf{Y}_{i})|1<i<N\}$, and its corresponding
feature space and label space are $\mathbf{X}\in R^{d}$ and $\mathbf{Y}=\{0,1\}^{q}$
where $q$ is the number of candidate labels, respectively. Given
testing datasets containing ID samples and OOD samples simultaneously,
the task of OODPLL is to induce a multi-class classifier $f:\mathbf{X}\rightarrow\mathbf{Y}$,
which can effectively detect the OOD samples and precisely predict
the ground-truth label of the unseen ID samples.

The PLOOD framework effectively overcomes the limitations of conventional PLL methods, which generally struggle with OOD samples. The PLOOD framework incorporates a novel \textbf{Positive-Negative Sample Augmentation (PNSA)} module that utilizes Positive-Unlabeled (PU) learning \cite{elkan2008learning} to generate strategically augmented positive and negative samples, thereby improving the model's capacity to differentiate OOD instances. Furthermore, label confidences are dynamically adjusted through features acquired from the PNSA module, addressing the label ambiguity present in PLL. Additionally, a newly developed \textbf{Partial Energy (PE)} OOD detection score effectively combines calibrated label confidences with an energy-based measure, thereby minimizing errors arising from ambiguous supervision. These contributions enable PLOOD to effectively unravel challenges in practical OOD PLL scenarios.

\subsection{PNSA Module: Adaptive Perturbation-Aware Feature Learning for OOD
Differentiation}

We introduce the PNSA module to effectively differentiate between in-distribution (ID) and OOD samples in PLL. PNSA synthesizes positive and negative augmented features adaptively, which explicitly shapes discriminative feature boundaries and improves the model's effectiveness in detecting OOD instances.

Given a training dataset  $\mathbf{D}=\{(\mathbf{x}_{i},\mathbf{Y}_{i})_{i=1}^{N}\}$, where each instance $\mathbf{x}_{i}$ corresponds to a candidate label set. Each sample is initially encoded into a $d$-dimensional feature representation:
\begin{equation}
\mathbf{h}_{i}=f_{\theta}(\mathbf{x}_{i})\in\mathbb{R}^{d},
\end{equation}

where $f_{\theta}$ represents a deep neural network characterized by weights $\theta$, which transforms input data into a $d$-dimensional embedding space. The resulting embedding $\mathbf{h}_{i}$ holds important semantic and structural data needed to tell the difference between ID and OOD instances.

\textbf{Adaptive Positive Sample Generation.} Positive augmented samples
aim to enrich the local representation around each original instance,
preserving the intrinsic manifold structure, which is defined as: 
\begin{equation}
\mathbf{h}_{i}^{+}=\mathbf{h}_{i}+\alpha\cdot\frac{1}{K}\sum_{\mathbf{h}_{j}\in\mathcal{N}_{K}(\mathbf{h}_{i})}(\mathbf{h}_{j}-\mathbf{h}_{i})+\epsilon_{p},
\end{equation}

where $\alpha > 0$ is a hyperparameter that regulates the extent of augmentation. The notation $\mathcal{N}_{K}(\mathbf{h}_{i})$ denotes the collection of $K$-nearest neighbors of $\mathbf{h}_{i}$ in the embedding space. To make sure there is local diversity, the perturbation term $\epsilon_{p}\sim\mathcal{N}(0,\sigma_{p}^{2}I)$ adds Gaussian noise in a controlled way. Positive samples stay within the intrinsic manifold boundary thanks to adaptive aggregation of neighboring information. This makes semantic consistency stronger for identification.

\textbf{Adaptive Negative Sample Generation.} In contrast to positive
samples, negative augmented samples explicitly simulate OOD instances.
To systematically push these features away from the intrinsic manifold
of ID samples, we define the negative sample generation process as: 
\begin{equation}
\mathbf{h}_{i}^{-}=\mathbf{h}_{i}+\beta\cdot\frac{\mathbf{h}_{i}-\bar{\mathbf{h}}}{\|\mathbf{h}_{i}-\bar{\mathbf{h}}\|_{2}}+\epsilon_{n},
\end{equation}

where $\beta > \alpha$ imposes more intense perturbations, resulting in augmented samples being positioned further from the ID manifold. The global center of all ID feature representations is shown by $\bar{\mathbf{h}}=\frac{1}{N}\sum_{j=1}^{N}\mathbf{h}_{j}$. It acts as a point of reference in the embedding space. The noise term $\epsilon_{n}\sim\mathcal{N}(0,\sigma_{n}^{2}I)$, where $\sigma_{n}>\sigma_{p}$, guarantees a higher variance of perturbation.
As a result, negative samples accurately reflect situations where the data is not distributed, which helps to improve the limits of feature discrimination.

\textbf{Perturbation-Aware Contrastive Loss (PACL)}. To explicitly
guide the feature learning process towards effective discrimination
of ID and OOD representations, we introduce the PACL: 
\begin{equation}
\mathcal{L}_{\text{PACL}}=-\frac{1}{N}\sum_{i=1}^{N}\log\frac{\exp(\mathrm{sim}^{\mathbf{h}^{+}}/\tau)}{\exp(\mathrm{sim}^{\mathbf{h}^{+}}/\tau)+\exp(\mathrm{\mathrm{sim}^{\mathbf{h}^{-}}}/\tau)},
\end{equation}

where $\beta > \alpha$ imposes more intense perturbations to propel augmented samples further from the ID manifold.  In this context, $\bar{\mathbf{h}}=\frac{1}{N}\sum_{j=1}^{N}\mathbf{h}_{j}$ signifies the global centroid of all ID feature representations, serving as a global reference point within the embedding space.  The noise term $\epsilon_{n}\sim\mathcal{N}(0,\sigma_{n}^{2}I)$, where $\sigma_{n}>\sigma_{p}$, guarantees a higher perturbation variance.
Consequently, negative samples effectively replicate OOD scenarios, enhancing the precision of feature discrimination boundaries.

\subsection{Label Disambiguation via Confidence Calibration using PNSA Module}

PLOOD seeks to effectively disambiguate partial labels using the feature representation derived from the PNSA module.  For an instance \( \mathbf{x}_{i} \) with a candidate label set \( \mathbf{Y}_{i}=\{\mathbf{y}_{i,j}\}_{j=1}^{q} \), we initially compute the predictive probability distribution over labels utilizing the calibrated label confidence.  The likelihood of assigning label $\mathbf{y}_{j}$ to instance $\mathbf{x}_{i}$ is expressed as:
\begin{equation}
P(\mathbf{y}_{j}|\mathbf{x}_{i})=\frac{\exp(\mathbf{h}_{i}^{\top}w_{j})}{\sum_{\mathbf{y}_{m}\in \mathbf{Y}_{i}}\exp(\mathbf{h}_{i}^{\top}w_{m})},
\end{equation}

where \( w_{j} \in \mathbb{R}^{d} \) represents the label embedding parameter associated with label \( \mathbf{y}_{j} \), and \( \mathbf{h}_{i} \) denotes the PNSA-based feature embedding for instance \( \mathbf{x}_{i} \).

The label confidence matrix $\mathcal{C}(\mathbf{y}_{i};\mathbf{y}_{j})$, initially set to uniform values, is updated dynamically during the training process through an iterative confidence calibration:
\begin{equation}
\mathcal{C}^{(t+1)}(\mathbf{x}_{i};\mathbf{y}_{j})=\mathcal{C}^{(t)}(\mathbf{x}_{i};\mathbf{y}_{j})\cdot P(\mathbf{y}_{j}|\mathbf{x}_{i}),
\end{equation}

followed by normalization across all candidate labels $\mathbf{Y}_{i}$ for
instance $\mathbf{x}_{i}$: 
\begin{equation}
\mathcal{C}^{(t+1)}(\mathbf{x}_{i};\mathbf{y}_{j})\leftarrow\frac{\mathcal{C}^{(t+1)}(\mathbf{x}_{i};\mathbf{y}_{j})}{\sum_{\mathbf{y}_{m}\in \mathbf{Y}_{i}}\mathcal{C}^{(t+1)}(\mathbf{x}_{i};\mathbf{y}_{m})}.
\end{equation}

The final partial label loss function, which incorporates the dynamically calibrated label confidence alongside the Perturbation-Aware Contrastive Loss (PACL) derived from the PNSA module, is expressed as follows: : 
\begin{equation}
\mathcal{L}_{final}=\underbrace{-\frac{1}{N}\sum_{i=1}^{N}\sum_{\mathbf{y}_{j}\in \mathbf{Y}_{i}}\mathcal{C}(\mathbf{x}_{i};\mathbf{y}_{j})\log P(\mathbf{y}_{j}|\mathbf{x}_{i})}_{\text{Confidence-Calibrated PLL Loss}}+\lambda\underbrace{\mathcal{L}_{PACL}}_{\text{PNSA Loss}},
\end{equation}

where $\lambda>0$ serves to balance the disambiguation of partial labels and the enhancement of OOD discrimination.

The optimization of this unified loss guarantees that the model effectively achieves accurate label disambiguation through confidence calibration and robust OOD differentiation via the PNSA feature learning module.

\subsection{Partial-Energy Score-based OOD Detection Integrated with Label Confidence}

Based on the feature representations derived from the PNSA module and the calibrated label confidence scores, we propose a new Partial-Energy (PE) Score method to enhance the efficacy of OOD detection in PLL. 

\textbf{Partial-Energy (PE) Formulation.}  We propose leveraging an energy-based function that integrates label confidence calibration, facilitating simultaneous OOD detection and partial-label disambiguation.  For a given instance \( \mathbf{x}_{i} \), the partial-energy score is formally defined as: 
\begin{equation}
\mathcal{E}_{\text{PE}}(\mathbf{x}_{i})=-\log\sum_{\mathbf{y}_{j}\in \mathbf{Y}_{i}}\mathcal{C}(\mathbf{x}_{i};\mathbf{y}_{j})\exp(\mathbf{h}_{i}^{\top}w_{j}),
\end{equation}

where $\mathbf{h}_{i}$ is the PNSA-based feature representation of $\mathbf{x}_{i}$,
$w_{j}$ is the embedding vector of candidate label $\mathbf{y}_{j}$, and
$\mathcal{C}(\mathbf{x}_{i};\mathbf{y}_{j})$ is the calibrated confidence for label $\mathbf{y}_{j}$
computed in Eqs. (6)-(7).

Intuitively, lower $\mathcal{E}_{\text{PE}}(\mathbf{x}_{i})$ values indicate that the
sample is closer to the learned distribution (i.e., likely an ID sample),
while higher values suggest that the sample significantly deviates
from the learned distribution, hence is more likely to be OOD.

From a probabilistic perspective, the PE score $\mathcal{E}_{\text{PE}}(\mathbf{x}_{i})$
can be related to the posterior predictive distribution over candidate
labels: 
\begin{align}
p(\mathbf{y}_{i}|\mathbf{Y}_{i}) & \propto\sum_{\mathbf{y}_{j}\in \mathbf{Y}_{i}}p(\mathbf{y}_{j}|\mathbf{x}_{i})p(\mathbf{x}_{i}|\mathbf{y}_{j})\\
 & \approx\sum_{\mathbf{y}_{j}\in \mathbf{Y}_{i}}\mathcal{C}(\mathbf{x}_{i};\mathbf{y}_{j})\exp(\mathbf{h}_{i}^{\top}w_{j}).\nonumber 
\end{align}

The partial-energy score quantifies the negative log-likelihood of observing instance $\mathbf{x}_{i}$ within the calibrated PLL distribution. Therefore, reducing the PE score is equivalent to enhancing the likelihood of ID samples while effectively distinguishing OOD samples through their elevated PE scores.

\textbf{Aggregation of Confidence for Global Calibration.} To enhance
robustness in the estimation of PE scores, we further define the Global
Label Confidence ($\mathcal{G}_{LC}$) for candidate label $\mathbf{y}_{j}$ as a global
calibration factor: 
\begin{equation}
\mathcal{G}_{LC}(\mathbf{y}_{j})=\frac{\mu_{\mathbf{y}_{j}}}{\sigma_{\mathbf{y}_{j}}+\epsilon},
\end{equation}

where $\mu_{\mathbf{y}_{j}}$ and $\sigma_{\mathbf{y}_{j}}$ represent the mean and
standard deviation of the calibrated confidences for label $\mathbf{y}_{j}$
across all training samples, respectively, and $\epsilon>0$ is a
small constant to ensure numerical stability.

The adjusted Global PE Score incorporating $\mathcal{G}_{LC}$ becomes: 
\begin{equation}
\mathcal{E}_{\text{G-PE}}(\mathbf{x}_{i})=-\log\sum_{\mathbf{y}_{j}\in Y_{i}}\mathcal{G}_{LC}(\mathbf{y}_{j})\mathcal{C}(\mathbf{x}_{i};\mathbf{y}_{j})\exp(\mathbf{h}_{i}^{\top}w_{j}).
\end{equation}

By incorporating $\mathcal{G}_{LC}$, the PE score is globally calibrated, thereby
enhancing consistency in OOD detection.

\textbf{Final OOD Detection Criterion.} Ultimately, we classify an
instance $\mathbf{x}_{i}$ as OOD if its Global Partial-Energy score exceeds
a predefined threshold $\gamma$: 
\begin{equation}
\text{OOD Decision}(\mathbf{x}_{i})=\mathbf{1}[\mathcal{E}_{\text{G-PE}}(\mathbf{x}_{i})>\gamma].
\end{equation}

The threshold $\gamma$ is empirically established on a held-out validation set to optimize detection performance. Our proposed PE-based method integrates label confidences and discriminative features, resulting in enhanced and interpretable out-of-distribution detection capabilities within the PLL framework.

\begin{algorithm}[htbp]
\caption{PLOOD}
\label{alg:PLOOD}
\KwIn{Dataset $\mathbf{D}=\{(\mathbf{x}_i, \mathbf{Y}_i)\}_{i=1}^{N}$, epochs $T$, augmentation count $M$, hyperparameters $\lambda, \beta, \gamma$.}
\KwOut{Optimized model $\theta^*$, label confidence $\mathcal{C}$.}

\textbf{Training Phase:} 

\textbf{Initialize} model $\theta$, confidence matrix $\mathcal{C}$.

\For{epoch $= 1, 2, \dots, T$}{
    \For{batch $\{(\mathbf{x}_i, \mathbf{Y}_i)\}_{i=1}^{B}$}{
        Generate positive and  negative augmented samples.

        Extract features: $\mathbf{h}_i, \mathbf{h}_i^+, \mathbf{h}_{i,m}^-$.

        Compute PACL loss $\mathcal{L}_{\text{PACL}}$ (Eq.4).

        Compute $P_{i,j}$ for each candidate label (Eq.5).

        Update label confidence matrix $\mathcal{C}$ (Eq.7).

        Compute PLL loss $\mathcal{L}_{\text{PLL}}$ (Eq.8).
        
        Optimize $\mathcal{L}(\theta) = \mathcal{L}_{\text{PLL}} + \lambda \mathcal{L}_{\text{PACL}}$ (Eq.8).

        Update $\theta \leftarrow \theta - \gamma \nabla_{\theta} \mathcal{L}(\theta)$.
    }
}

\textbf{Inference Phase:} 

\For{each test sample $\mathbf{x}_i$}{
    Compute $h_i = f_{\theta}(\mathbf{x}_i)$.

    Compute global label confidence $\mathcal{G}_{LC}$ (Eq.11).

    Compute PE score $\mathcal{E}_{\text{PE}}(x_i)$ (Eq.12).

    \If{$\mathcal{E}_{\text{PE}}(x_i) > \gamma$}{
        Mark $\mathbf{x}_i$ as OOD.
    }
    \Else{
        Predict $\arg\max_{\mathbf{y}_j} P_{i,j}$.
    }
}

\Return $\theta^*$ and updated $\mathcal{C}$.

\end{algorithm}

\section{Experiments}

This section presents a comprehensive
overview of the experimental setup, the results, and the analysis
and insights gleaned from the comparative studies.

\subsection{Experimental Setup}

\textbf{Dataset Selection:} The selection of datasets is essential for evaluating the robustness and adaptability of the PLOOD framework in differentiating between ID and OOD data.
\begin{itemize} 
\item \textbf{ID datasets:} We assess the framework's performance using CIFAR-10 and CIFAR-100 as the identification datasets. CIFAR-10 comprises 60,000 images across ten categories to evaluate the model's generalization capabilities, whereas CIFAR-100 offers 100 categories to further assess the model's differentiation abilities.
\item \textbf{OOD Datasets:} Our framework is tested on several datasets for Out-of-Distribution (OOD) detection. These include FashionMNIST, the Describable Textures Dataset (DTD), LSUNCrop, TinyImageNetCrop, and Fooling Images. These datasets encompass a range of domains, including environmental, architectural, and object scenes. The datasets exhibit distinct complexities and intricate classification tasks, including environmental, architectural, and object scenes.
\item \textbf{Practical PLL datasets}: This study evaluates five real-world PLL datasets from various domains: Lost (face naming from videos with face crops and caption names), MSRCv2 (object classification with image segments and co-occurring objects), BirdSong (bird song classification with syllables and concurrent species), Soccer Player, and Yahoo! News (face naming from images with face crops and subtitle names).
\end{itemize}

\textbf{Comparing Method:} We selected the SOTA PLL algorithms PaPi \cite{xia2023towards}, CSDLE \cite{he2023candidate}, CroSel \cite{tian2024crosel}, SoDisam \cite{jiang2024navigating}, and LS-PLL \cite{gong2024does} to evaluate PLL using OOD detection techniques in a fair and effective manner.
The comparative PLL methods consist of deep-based PLL models utilizing a guided prototypical classifier. The detection methods included are Entropy \cite{chan2021entropy}, ODIN \cite{liang2017enhancing}, DML \cite{zhang2023decoupling}, EnergyBased \cite{liu2020energy}, and JointEnergy (J.Energy) \cite{wang2021can}. 

\textbf{Evaluation Metrics}: Two well-known OOD metrics were used to judge our method: Area Under the Precision-Recall Curve for In-distribution (AUPR-IN) and False Positive Rate at 95\% True Positive Rate (FPR95).

\textbf{Training Details:} All the PLL methods utilize the ResNet34 architecture as the backbone. The training process begins with a learning rate of 0.001, gradually altering it in accordance with a decay schedule. We set the batch size at 128 and select the Adam optimizer over 100 epochs. The experiments were performed on a Windows system equipped with a 13th Gen Intel(R) Core(TM) i7-13700K processor at 3.40 GHz, an RTX 4080 GPU, 32GB of RAM, and utilizing PyTorch environments.

\subsection{Experimental Results }

\subsubsection*{Whether OOD scenario is necessity for PLL? }

We conducted a comparative study on the CIFAR-10 dataset to investigate the performance of PLL in OOD scenarios and to emphasize the importance of OODPLL. This study evaluated several SOTA PLL methods, including PaPi \cite{xia2023towards}, CSDLE \cite{he2023candidate}, CroSel \cite{tian2024crosel}, SoDisam \cite{jiang2024navigating}, and LS-PLL \cite{gong2024does}, in comparison to the proposed PLOOD framework.The evaluation was conducted in two distinct settings: closed-set conditions, in which all test samples adhered to the same distribution as the training set, and open-set conditions, where 20–30\% of OOD samples from a disjoint dataset were incorporated into the CIFAR-10 test set to replicate real-world distribution shifts. The evaluation metric employed to assess the robustness of each method in dealing OOD objects in PLL was overall accuracy. The findings are presented in the table below:
\begin{table}[tbh]
\caption{Accuracy(\%) of six SOTA PLL methods on CIFAR-10 dataset under ID
(No-OOD) and OOD (With-OOD) scenarios.}

\begin{centering}
\begin{tabular}{rcc}
\hline 
Method & Acc.(No-OOD) & Acc.(With-OOD)\tabularnewline
\hline 
PaPi \cite{xia2023towards}  & 96.90 & 75.3\tabularnewline
CSDLE \cite{he2023candidate} & 96.28 & 73.7\tabularnewline
CroSel \cite{tian2024crosel} & \textbf{97.34} & 74.1\tabularnewline
SoDisam \cite{jiang2024navigating} & 96.64 & 72.9\tabularnewline
LS-PLL \cite{gong2024does} & 95.97  & 72.5\tabularnewline
\hline 
\textbf{PLOOD } & 97.02 & \textbf{87.5}\tabularnewline
\hline 
\end{tabular}
\par\end{centering}
\end{table}

\begin{table*}[tbh]
\caption{Performance comparisons of the \textbf{PLOOD} framework and \textbf{two
existing PLL methods with five OOD scores} on the \textbf{five OOD
datasets} of \textbf{CIFAR-10} under partial rate ($p=0.1$). \textbf{AUPR-IN}(\%)
and \textbf{FPR95}(\%) are used for the evaluation.}

\begin{spacing}{1.2}
\centering{}%
\begin{tabular}{r|r|c|c|c|c|c}
\hline 
\multicolumn{2}{c|}{{\small{}Metrics}} & \multicolumn{5}{c}{{\small{}AUPR-IN ($\uparrow$) / FPR95 ($\downarrow$)}}\tabularnewline
\hline 
\multirow{1}{*}{PLL methods} & OOD scores & {\small{}Fashion} & {\small{}LSUN} & {\small{}Tiny} & {\small{}DTD} & {\small{}Fooling}\tabularnewline
\hline 
\multirow{5}{*}{PaPi \cite{xia2023towards}} & {\small{}ODIN} \cite{liang2017enhancing} & {\small{}56.20/86.37} & {\small{}67.95/80.55} & {\small{}55.13/91.62} & {\small{}53.95/93.09} & {\small{}67.10/89.31}\tabularnewline
 & {\small{}DML} \cite{zhang2023decoupling} & {\small{}66.05/77.10} & {\small{}72.58/76.06} & {\small{}68.46/82.17} & {\small{}53.90/90.77} & {\small{}56.74/95.69}\tabularnewline
 & {\small{}Entropy} \cite{chan2021entropy} & {\small{}\uline{70.41}}{\small{}/}{\small{}\uline{74.79}} & {\small{}67.39/79.69} & {\small{}64.47/84.35} & {\small{}52.67/92.18} & {\small{}65.41/88.68}\tabularnewline
 & {\small{}Energy} \cite{liu2020energy} & {\small{}67.14/77.12} & {\small{}75.08/73.76} & {\small{}70.86/82.21} & {\small{}\uline{55.09}}{\small{}/}{\small{}\uline{88.31}} & {\small{}67.15/85.32}\tabularnewline
 & {\small{}J.Energy} \cite{wang2021can} & {\small{}68.98/75.48} & {\small{}\uline{76.65}}{\small{}/}{\small{}\uline{72.08}} & {\small{}\uline{71.50}}{\small{}/}{\small{}\uline{79.04}} & {\small{}54.11/89.01} & {\small{}\uline{67.31}}{\small{}/}{\small{}\uline{85.01}}\tabularnewline
\hline 
\multirow{5}{*}{CroSel \cite{tian2024crosel}} & {\small{}ODIN} \cite{liang2017enhancing} & {\small{}51.57/88.67} & {\small{}70.57/88.20} & {\small{}56.07/92.98} & {\small{}45.96/97.86} & {\small{}62.61/89.61}\tabularnewline
 & {\small{}DML} \cite{zhang2023decoupling} & {\small{}65.21/82.46} & {\small{}69.19/89.98} & {\small{}65.81/84.52} & {\small{}50.53/93.10} & {\small{}61.91/90.18}\tabularnewline
 & {\small{}Entropy} \cite{chan2021entropy} & {\small{}65.13/81.81} & {\small{}70.22/85.54} & {\small{}65.86/83.88} & {\small{}51.01/91.46} & {\small{}62.48/89.47}\tabularnewline
 & {\small{}Energy} \cite{liu2020energy} & {\small{}\uline{65.38}}{\small{}/}{\small{}\uline{80.46}} & {\small{}\uline{74.47}}{\small{}/}{\small{}\uline{79.01}} & {\small{}\uline{65.97}}{\small{}/}{\small{}\uline{83.77}} & {\small{}54.69/88.10} & {\small{}62.94/90.18}\tabularnewline
 & {\small{}J.Energy} \cite{wang2021can} & {\small{}62.64/85.31} & {\small{}70.20/82.79} & {\small{}62.74/90.42} & {\small{}\uline{55.39}}{\small{}/}{\small{}\uline{87.79}} & {\small{}\uline{64.60}}{\small{}/}{\small{}\uline{89.28}}\tabularnewline
\hline 
\multicolumn{2}{c|}{\textbf{\small{}PLOOD}} & \textbf{\small{}76.60/69.93} & \textbf{\small{}82.98/62.40} & \textbf{\small{}79.08/69.17} & \textbf{\small{}64.53/80.46} & \textbf{\small{}77.84/73.20}\tabularnewline
\hline 
\end{tabular}
\end{spacing}
\end{table*}
\begin{table*}[tbh]
\caption{Performance comparisons of the \textbf{PLOOD} framework and \textbf{two
existing PLL methods with five OOD scores} on the \textbf{five OOD
datasets} of \textbf{CIFAR-100} under partial rate ($p=0.1$). \textbf{AUPR-IN}(\%)
and \textbf{FPR95}(\%) are used for the evaluation.}

\begin{spacing}{1.2}
\centering{}%
\begin{tabular}{r|r|c|c|c|c|c}
\hline 
\multicolumn{2}{c|}{{\small{}Metrics}} & \multicolumn{5}{c}{{\small{}AUPR-IN ($\uparrow$) / FPR95 ($\downarrow$)}}\tabularnewline
\hline 
\multirow{1}{*}{PLL methods} & OOD scores & {\small{}Fashion} & {\small{}LSUN} & {\small{}Tiny} & {\small{}DTD} & {\small{}Fooling}\tabularnewline
\hline 
\multirow{5}{*}{PaPi \cite{xia2023towards}} & {\small{}ODIN} \cite{liang2017enhancing} & {\small{}62.75/82.98} & {\small{}63.17/92.21} & {\small{}55.23/93.09} & {\small{}43.83/88.09} & {\small{}51.22/92.08}\tabularnewline
 & {\small{}DML} \cite{zhang2023decoupling} & {\small{}60.18/84.61} & {\small{}62.57/77.96} & {\small{}63.97/80.44} & {\small{}41.52/92.46} & {\small{}50.55/96.86}\tabularnewline
 & {\small{}Entropy} \cite{chan2021entropy} & {\small{}64.19/78.78} & {\small{}63.56/81.12} & {\small{}63.20/79.94} & {\small{}41.76/90.15} & {\small{}49.78/98.18}\tabularnewline
 & {\small{}Energy} \cite{liu2020energy} & {\small{}63.96/79.98} & {\small{}64.18/77.76} & {\small{}62.55/78.66} & {\small{}43.11/89.29} & {\small{}\uline{52.43}}{\small{}/}{\small{}\uline{88.85}}\tabularnewline
 & {\small{}J.Energy} \cite{wang2021can} & {\small{}\uline{64.69}}{\small{}/}{\small{}\uline{78.08}} & {\small{}\uline{65.30}}{\small{}/}{\small{}\uline{75.64}} & {\small{}\uline{65.16}}{\small{}/}{\small{}\uline{76.20}} & {\small{}\uline{44.24}}{\small{}/}{\small{}\uline{87.45}} & {\small{}52.19/90.16}\tabularnewline
\hline 
\multirow{5}{*}{CroSel \cite{tian2024crosel}} & {\small{}ODIN} \cite{liang2017enhancing} & {\small{}62.69/82.49} & {\small{}61.79/86.08} & {\small{}53.32/89.93} & {\small{}41.95/90.93} & {\small{}54.37/89.09}\tabularnewline
 & {\small{}DML} \cite{zhang2023decoupling} & {\small{}62.72/80.68} & {\small{}62.97/80.38} & {\small{}59.97/82.77} & {\small{}35.83/94.89} & {\small{}53.27/88.56}\tabularnewline
 & {\small{}Entropy} \cite{chan2021entropy} & {\small{}60.31/81.66} & {\small{}61.91/84.30} & {\small{}58.57/83.18} & {\small{}35.35/95.78} & {\small{}52.29/89.26}\tabularnewline
 & {\small{}Energy} \cite{liu2020energy} & {\small{}64.53/79.79} & {\small{}\uline{64.41}}{\small{}/}{\small{}\uline{80.01}} & {\small{}57.80/82.77} & {\small{}43.29/89.42} & {\small{}54.30/88.17}\tabularnewline
 & {\small{}J.Energy} \cite{wang2021can} & {\small{}\uline{65.54}}{\small{}/}{\small{}\uline{77.82}} & {\small{}60.84/83.06} & {\small{}\uline{61.07}}{\small{}/}{\small{}\uline{80.23}} & {\small{}\uline{46.03}}{\small{}/}{\small{}\uline{87.59}} & {\small{}\uline{54.76}}{\small{}/}{\small{}\uline{87.41}}\tabularnewline
\hline 
\multicolumn{2}{c|}{\textbf{\small{}PLOOD}} & \textbf{\small{}73.49/68.33} & \textbf{\small{}72.38/69.62} & \textbf{\small{}71.31/71.36} & \textbf{\small{}51.46/81.43} & \textbf{\small{}61.71/80.42}\tabularnewline
\hline 
\end{tabular}
\end{spacing}
\end{table*}

The results indicate that all methods attain high accuracy in closed-set conditions (between 95.97\% and 97.34\%), yet their performance significantly decreased when faced with OOD samples. Baseline methods, for instance, exhibit a marked decline of roughly 24–27 percentage points, as seen with PaPi, which drops from 96.90\% to 72.3\%.  In contrast, PLOOD demonstrates strong performance, with a modest decline of approximately 9.5 percentage points (from 97.02\% to 87.5\%).  This notable difference underscores the importance of integrating OOD detection into PLL methods.  PLOOD's integration of PNSA feature learning and PE-based label refinement significantly improves its generalization capabilities in open-set scenarios.

\subsubsection*{PLOOD achieves optimal performance compared to existing SOTA PLL
methods with ODD detectors}

The results indicate that OODPLL represents an emerging research direction, without  no dedicated SOTA model specifically designed for this task presently. We conducted experiments on CIFAR-10 and CIFAR-100 with a partial labeling rate of $(p=0.1)$ to evaluate the effectiveness of PLOOD, comparing it against two prominent PLL methods: PaPi and CroSel. Each baseline method was combined with five commonly utilized  OOD scoring techniques: ODIN, DML, Entropy, Energy, and J.Energy. The models underwent evaluation across five OOD datasets: Fashion, LSUN, Tiny, DTD, and Fooling. Two primary metrics were employed: AUPR-IN (where higher values indicate better performance), and FPR95 (where lower values are preferable).

The findings indicate that PLOOD significantly outperforms current PLL methods. On CIFAR-10, PaPi attains AUPR-IN scores ranging from 55.13\% to 70.41\%, whereas CroSel's newly generated results vary from 56.50\% to 69.50\%. PLOOD consistently outperforms both methods, achieving AUPR-IN values ranging from 64.53\% to 83.50\%, while also demonstrating significantly lower FPR95 values. On CIFAR-100, PaPi achieves AUPR-IN values ranging from 60.18\% to 64.69\%, while CroSel exhibits marginally better performance. However, both methods do not reach the superior performance of PLOOD, which ranges from 71.31\% to 73.49\%.

The results show a big problem with traditional PLL methods: they can't do accurate ID classification in OOD situations, even when standard OOD detection scores are added. PLOOD’s incorporation of PNSA and PE scoring mechanisms facilitates enhanced OOD identification and label disambiguation, thereby markedly improving robustness and generalization. PLOOD sets a new standard for OOD aware probabilistic label learning by dealing with label ambiguity and distribution shifts. This makes the method more reliable and adaptable for real-world open-set learning situations. In closed-set conditions, all methods are very accurate (ranging from 95.97\% to 97.34\%), but when they are used with OOD samples, they are much less accurate. Baseline methods exhibit a marked decline of approximately 24–27 percentage points (e.g., PaPi decreases from 96.90\% to 72.3\%). In contrast, PLOOD demonstrates strong performance, with a slight decline of approximately 9.5 percentage points (from 97.02\% to 87.5\%). This notable difference underscores the importance of integrating OOD detection into PLL methods.

\subsubsection*{Ablation studies of all the components contribute to PLOOD }

We conducted ablation experiments on CIFAR-10 with a partial labeling rate of $p=0.1$ to evaluate the contributions of the PNSA and PE modules within the PLOOD framework.  Models were evaluated on five disjoint datasets—Fashion, LSUN, Tiny, DTD, and Fooling—to simulate OOD scenarios, employing AUPR-IN as the exclusive metric for ID recognition.  The table below presents a summary of the results from the ablation study:

\begin{table}[tbh]
\caption{Ablation studies of all the components contribute to PLOOD ({\small{}\FiveStarOpen }:
PLOOD, \textbf{N}: PNSA, \textbf{P}: PE score, \textbf{C}: CSDLE, \textbf{S:} SoDisam, \textbf{L}: LS-PLL
, \textbf{O}: ODIN, \textbf{E}: Energy, \textbf{J}: J.Energy)}

\centering{}%
\begin{tabular}{llllll}
\hline 
{\small{}Methods} & {\small{}FA.} & {\small{}L.} & {\small{}T.} & {\small{}DTD} & FO.\tabularnewline
\hline 
\textbf{\small{}PLOOD} & \textbf{\small{}76.60} & \textbf{\small{}82.98} & \textbf{\small{}79.08} & \textbf{\small{}64.53} & \textbf{\small{}77.84}\tabularnewline
\hline 
{\small{}\FiveStarOpen{} w/o }\textbf{\small{}N}{\small{}+}\textbf{\small{}P} & {\small{}34.27} & {\small{}42.52} & {\small{}38.49} & {\small{}34.91} & {\small{}40.93}\tabularnewline
{\small{}\FiveStarOpen{} w/o }\textbf{\small{}N} & {\small{}65.03} & {\small{}70.57} & {\small{}68.88} & {\small{}52.73} & {\small{}66.91}\tabularnewline
{\small{}\FiveStarOpen{} w/o }\textbf{\small{}P} & {\small{}48.14} & {\small{}46.39} & {\small{}42.86} & {\small{}37.64} & {\small{}48.72}\tabularnewline
\hline 
{\small{}\FiveStarOpen{} w/o }\textbf{\small{}N}{\small{}(+}\textbf{\small{}C}
\cite{he2023candidate}{\small{})} & {\small{}66.02 } & {\small{}72.33 } & {\small{}67.76 } & {\small{}53.09 } & {\small{}67.12}\tabularnewline
{\small{}\FiveStarOpen{} w/o }\textbf{\small{}N}{\small{}(+}\textbf{\small{}S}{\small{}
}\cite{jiang2024navigating}{\small{})} & {\small{}63.10 } & {\small{}71.06} & {\small{}62.45} & {\small{}51.12 } & {\small{}61.12}\tabularnewline
{\small{}\FiveStarOpen{} w/o }\textbf{\small{}N}{\small{}(+}\textbf{\small{}L}{\small{}
}\cite{gong2024does}{\small{})} & {\small{}60.33} & {\small{}66.48} & {\small{}59.91} & {\small{}49.76} & {\small{}62.12}\tabularnewline
\hline 
{\small{}\FiveStarOpen{} w/o }\textbf{\small{}P}{\small{}(+}\textbf{O}{\small{}
}\cite{liang2017enhancing}{\small{})} & {\small{}62.11 } & {\small{}67.15} & {\small{}63.15} & {\small{}47.56} & {\small{}62.03}\tabularnewline
{\small{}\FiveStarOpen{} w/o }\textbf{\small{}P}{\small{}(+}\textbf{\small{}E}{\small{}
}\cite{liu2020energy}{\small{})} & {\small{}64.52 } & {\small{}63.88} & {\small{}59.47 } & {\small{}45.20} & {\small{}55.01}\tabularnewline
{\small{}\FiveStarOpen{} w/o }\textbf{\small{}P}{\small{}(+}\textbf{\small{}J}{\small{}
}\cite{wang2021can}{\small{})} & {\small{}65.44 } & {\small{}65.09} & {\small{}61.85} & {\small{}46.02} & {\small{}57.76}\tabularnewline
\hline 
\end{tabular}
\end{table}

The results indicate that PLOOD attains the highest AUPR-IN across all OOD datasets, with scores of 76.60 on Fashion and 82.98 on LSUN, thereby confirming its superior capacity to manage OOD instances.  The removal of both PNSA and PE modules (without \textbf{N+P}) results in a significant decrease in AUPR-IN (e.g., 34.27 on Fashion), underscoring their critical importance in feature representation and out-of-distribution detection. The removal of PE alone, without P, leads to a significant decrease in performance, as evidenced by the AUPR-IN dropping from 76.60 to 48.14 on Fashion. This demonstrates the importance of energy-based refinement in sustaining high in-distribution confidence.  Replacing PE with conventional OOD detection scores (ODIN, Energy, J.Energy) results in performance that is significantly lower than PLOOD, thereby confirming the effectiveness of PE in addressing label ambiguity.  Removing PNSA (without \textbf{N}) leads to a significant decrease in accuracy; however, it still outperforms the scenario without P. This suggests that PNSA improves ID representation but is inadequate for OOD detection in the absence of PE.  Replacing PNSA with current PLL methods (CSDLE, SoDisam, LS-PLL) does not achieve the performance level of PLOOD, demonstrating that PNSA distinctly improves feature learning and OOD differentiation.  The findings confirm that both PNSA and PE are essential, allowing PLOOD to effectively distinguish between ID and OOD samples, surpass existing methods, and set a new benchmark for OOD-aware PLL in open-set environments.

\subsubsection*{Practical Partial label dataset with OOD objects}

We evaluated the effectiveness of PLOOD by testing it on five real-world PLL datasets: MSRCV2, Lost, BirdSong, Soccer Player, and Yahoo. We tested PLOOD against five other advanced PLL methods—PaPi, CSDLE, CroSel, SoDisam, and LS-PLL—in the same conditions, using classification accuracy as the only measure. PLOOD attained scores of 55.60\%, 82.55\%, 76.15\%, 60.88\%, and 71.22\% on MSRCV2, Lost, BirdSong, Soccer Player, and Yahoo, respectively, surpassing all baseline models by approximately 10–15 percentage points. PLOOD on Lost exceeds the best baseline, CroSel, by more than 11 percentage points (82.55\% compared to 71.25\%). PLOOD demonstrates superior performance over the baselines on MSRCV2, exceeding them by 8 to 11 percentage points. The results show that current PLL methods struggle in open-set situations where samples can be OOD. PLOOD's enhanced performance substantiates its PNSA and PE mechanisms, which improve feature PLL, demonstrating robust adaptability and strong generalization.

\begin{figure}[tbh]
\centering \includegraphics[scale=0.63]{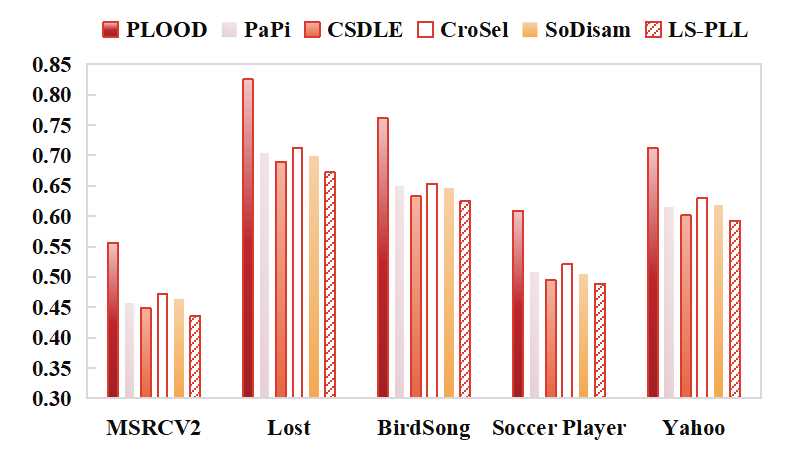} \caption{Accuracy of six SOTA PLL methods on five practical PLL datasets with
OOD objects.}
\end{figure}

\subsubsection*{Convergence Analysis Of PLOOD On CIFAR-100}

This study assesses the convergence of PLOOD through an experiment utilizing the CIFAR-100 dataset, monitoring the progression of total loss $\mathcal{L}(\theta)$ across training epochs. The total loss guarantees that PLOOD concurrently enhances label confidence, develops robust feature representations, and advances out-of-distribution detection. The model underwent training for 200 epochs, utilizing hyperparameters that were determined to be optimal based on sensitivity analysis. The convergence results show that $\mathcal{L}(\theta)$ steadily decreases and levels off around epoch 120. This suggests that PLOOD optimizes well without becoming unstable. Initially, $\mathcal{L}(\theta)$ decreases rapidly, indicating efficient early-stage learning of ID and OOD representations. The loss stops changing after 120 iterations, which means that the model has reached convergence as it improves its decision boundaries and label confidence estimation. The stability of $\mathcal{L}(\theta)$ in subsequent epochs indicates that PLOOD is effectively optimized, mitigates overfitting, and attains robust convergence in open-set PLL tasks.
\begin{figure}[tbh]
\centering{}\centering \includegraphics[scale=0.5]{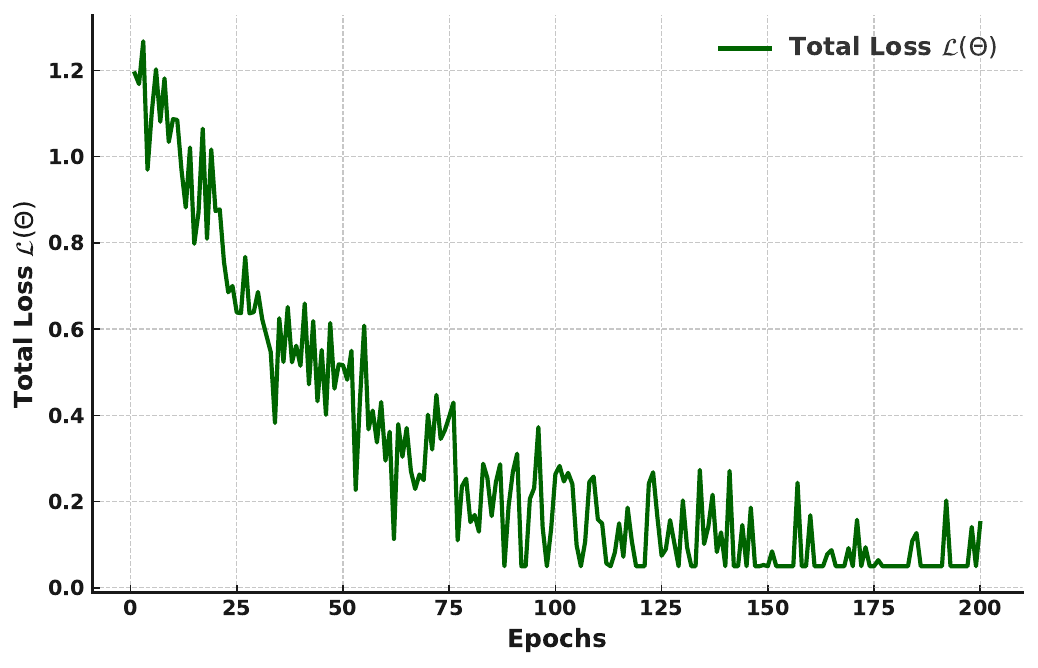}
\caption{Convergence Analysis of PLOOD On CIFAR-100.}
\end{figure}

\section{Conclusion}

This paper presents PLOOD, a new framework aimed at effectively tackling the OODPLL problem in open-set scenarios. PLOOD distinguishes itself from traditional PLL methods by incorporating a Positive-Negative Sample Augmentation (PNSA) module and a Partial Energy (PE)–based out-of-distribution (OOD) detection module, thereby significantly improving model robustness. The PNSA module generates both positive and negative augmented samples, facilitating distinct feature differentiation between ID and OOD instances. The PE module integrates dynamically calibrated label confidences within an energy-based detection framework, significantly reducing errors associated with ambiguous PLL supervision.Comprehensive experiments on the CIFAR-10 and CIFAR-100 datasets indicate that PLOOD consistently surpasses current SOTA methods in classification accuracy and OOD detection capability, highlighting the importance of incorporating OOD awareness into probabilistic label learning (PLL) models. PLOOD establishes a new benchmark for robust PLL in trustworthy conditions, providing a solid foundation for future research in open-set and real-world scenarios.

{
    \small
    \bibliographystyle{ieeenat_fullname}
    \bibliography{main}
}

\end{document}